\documentclass{article}

\usepackage{microtype}
\usepackage{graphicx}
\usepackage{subfigure}
\usepackage{booktabs} % for professional tables

\usepackage{soul}
\usepackage[utf8]{inputenc} % allow utf-8 input
\usepackage[T1]{fontenc}    % use 8-bit T1 fonts
\usepackage{hyperref}       % hyperlinks
\usepackage{url}            % simple URL typesetting
\usepackage{booktabs}       % professional-quality tables
\usepackage{amsfonts}       % blackboard math symbols
\usepackage{nicefrac}       % compact symbols for 1/2, etc.
\usepackage{microtype}      % microtypography
\usepackage{xcolor}         % colors
\usepackage{natbib}
\usepackage{graphicx,wrapfig,lipsum}
\usepackage{amsmath}
\usepackage{amsfonts}
\usepackage{multirow}
\usepackage{lscape}
\usepackage{float}
\usepackage{caption}
\newcommand{\betahat}{\mathbf{\hat{\beta}}}
\newcommand{\FIM}{\mathbf{F(\betahat})}
\newcommand{\FIMinv}{\mathbf{F^{-1}(\betahat)}}

\newenvironment{psmallmatrix}
  {\left(\begin{smallmatrix}}
  {\end{smallmatrix}\right)}

\usepackage{hyperref}

\usepackage[accepted]{icml2023}

\usepackage{amsmath}
\usepackage{amssymb}
\usepackage{mathtools}
\usepackage{amsthm}

\usepackage[capitalize,noabbrev]{cleveref}

\theoremstyle{plain}

\theoremstyle{definition}

\theoremstyle{remark}

\usepackage[textsize=tiny]{todonotes}

\icmltitlerunning{Structural Neural Additive Models}

\begin{document}

\twocolumn[
\icmltitle{Structural Neural Additive Models: \\
Enhanced Interpretable Machine Learning}




\begin{icmlauthorlist}
\icmlauthor{Mattias Luber}{yyy}
\icmlauthor{Anton Thielmann}{comp}
\icmlauthor{Benjamin Säfken}{comp}

\end{icmlauthorlist}

\icmlaffiliation{yyy}{Third Institute of Physics, Georg-August University, Goettingen, Germany}
\icmlaffiliation{comp}{Chair of Data Science and Applied Statistics, TU Clausthal
}

\icmlcorrespondingauthor{Mattias Luber}{mattias.luber@phys.uni-goettingen.de}

\icmlkeywords{Machine Learning, ICML}

\vskip 0.3in
]



%\printAffiliationsAndNotice{} % otherwise use the standard text.

%\maketitle

\begin{abstract}
Deep neural networks (DNNs) have shown exceptional performances in a wide range of tasks and have become the go-to method for problems requiring high-level predictive power. There has been extensive research on how DNNs arrive at their decisions, however, the inherently uninterpretable networks remain up to this day mostly unobservable "black boxes". In recent years, the field has seen a push towards interpretable neural networks, such as the visually interpretable Neural Additive Models (NAMs). We propose a further step into the direction of intelligibility beyond the mere visualization of feature effects and propose Structural Neural Additive Models (SNAMs). A modeling framework that combines classical and clearly interpretable statistical methods with the predictive power of neural applications. Our experiments validate the predictive performances of SNAMs. The proposed framework performs comparable to state-of-the-art fully connected DNNs and we show that SNAMs can even outperform NAMs while remaining inherently more interpretable.

\end{abstract}

\section{Introduction}
In the recent years deep neural networks (DNNs) turned out to be extremely effective in modelling difficult high dimensional problems and in recognizing complex patterns in unstructured data. While the reasons for their performance is yet not fully understood, it is undoubted that DNNs form the current state-of-the-art-approach in many disciplines such as computer vision, machine translation and speech recognition \cite{sejnowski_unreasonable_2020}. Their effectiveness however, comes at a price. Due to their "black box" nature, DNNs lack interpretability, which limits their application in data sensitive domains like health care, finance or insurance \cite{adadi_peeking_2018}. 
To address this limitations, several approaches have been developed based on post-hoc investigations of the model predictions. Common examples here are LIME \cite{ribeiro_why_2016} or layer wise relevance propagation (LRP) \cite{bach2015pixel}. Although those approaches might be able to indicate how individual predictions are generated, they do not provide a global and complete picture of the underlying decision making process. 
Neural Additive Models (NAMs) \citep{agarwal_neural_2021} were recently proposed as a class of Neural Networks, that impose an additivity constraint on the input data and thus  allow to directly derive the feature-wise contribution onto the generated predictions as a function of the input domain. While this indeed yields an exact representation of the decision making process, NAMs are nevertheless highly complex functions that are characterized by hundreds of thousands of parameters and thus fail to address additional dimensions of interpretability \cite{murdoch2019definitions}.
To this end, we propose the use of Structural Neural Additive Models (SNAMs) as a way to achieve the same (and even better) predictive performance with a fraction of the parameters required, while providing intelligibility beyond mere visualizations. The contributions of SNAMs can be summarized as follows:

\begin{itemize}
    \item  SNAMs enable direct interpretation of the estimated parameters, as well as a measure of model uncertainty based on Bayesian confidence intervals. 
    \item Parameter sparsity is achieved through spline-based activation layers that mimic the behaviour of cubic regression splines and their ability to model nonlinear functions with a minimum of parameters. 
    \item We propose fast and computationally efficient neural splines based on the Silverman kernel.
    \item Due to the extreme parameter sparsity, intelligibility beyond visualization is possible and quantification of parameter uncertainty and hence the construction of confidence bands is achieved.
    \item Learnable knots allow SNAMs to model highly non-linear functional relationships turning the spline based approach into a highly flexible model class, despite the extreme parameter sparsity.
    \item A special hallmark of the spline based activation layers is, that they directly generalize to higher dimensions and thereby form a native way to process and interpret spatial data. The effectiveness of this property is demonstrated on the Carlifornia Housing dataset along with an evaluation of the full interpretability of SNAMs.
    \item The predictive performance of SNAMs is evaluated with a benchmark study. We analyze multiple datasets and compare the results to multiple neural- as well as non-neural models.

    %In this work we first reformulate the basis expansion with cubic regression splines as a neural activation layer and then propose a fast and computationally efficient approximation based on the Silverman kernel. The predictive performance of SNAMs is benchmarked  against ordinary NAMs and other state-of-the-art models on four relevant datasets. A special hallmark of the spline based activation layers is, that they directly generalization to higher dimension and thereby form a native way to process and interpret spatial data. The effectiveness of this property is demonstrated on the Carlifornia Housing dataset along with an evaluation of the full interpretability of SNAMs.
\end{itemize}

The paper is structured as follows: First, we shortly summarize generalized additive models \citep{hastie_generalized_1986} and traditional smoothers. Subsequently, we reformulate the basis expansion with cubic regression splines as a neural activation layer and then propose a fast and computationally efficient approximation based on the Silverman kernel. Additionally, we introduce learnable knots and thus adapt neural splines to account for jagged shape functions. Section 3 demonstrates the interpretability of SNAMs via the construction of the Fisher Information Matrix. In section 4 we conduct the benchmark study and compare SNAMs to other state-of-the-art models. Section 5 concludes.

\section{Structural Neural Additive Models}\label{sec:cubic_spline_unit}

Generalized Additive Models (GAM) were developed by Hastie and Tibshirani \cite{hastie_generalized_1986} to model flexible nonlinear effects in regression and classification problems as an extension of the more restrictive Generalized Linear Models (GLM) \cite{nelder_generalized_1972}. 
In its fundamental form, a GAM can be expressed as

\begin{equation}
    \eta = \beta_0 + \sum_i^p f_i(x_{i}) +\sum_{i,j}^p f_{ij}(x_i,x_j) + \epsilon, \mu = g(\eta),
\end{equation}
where $\eta$ is used to model the expectation of the target variable $y$ given some features $x \in \mathbb{R}^{(n,p)}$ via link-function $g(\eta) =\mu \equiv  E(y|x)$ and $y\in \mathbb{R}^{(n,1)}$ is further assumed to follow a distribution of the exponential family.

\begin{figure}
\includegraphics[width=.5\textwidth]{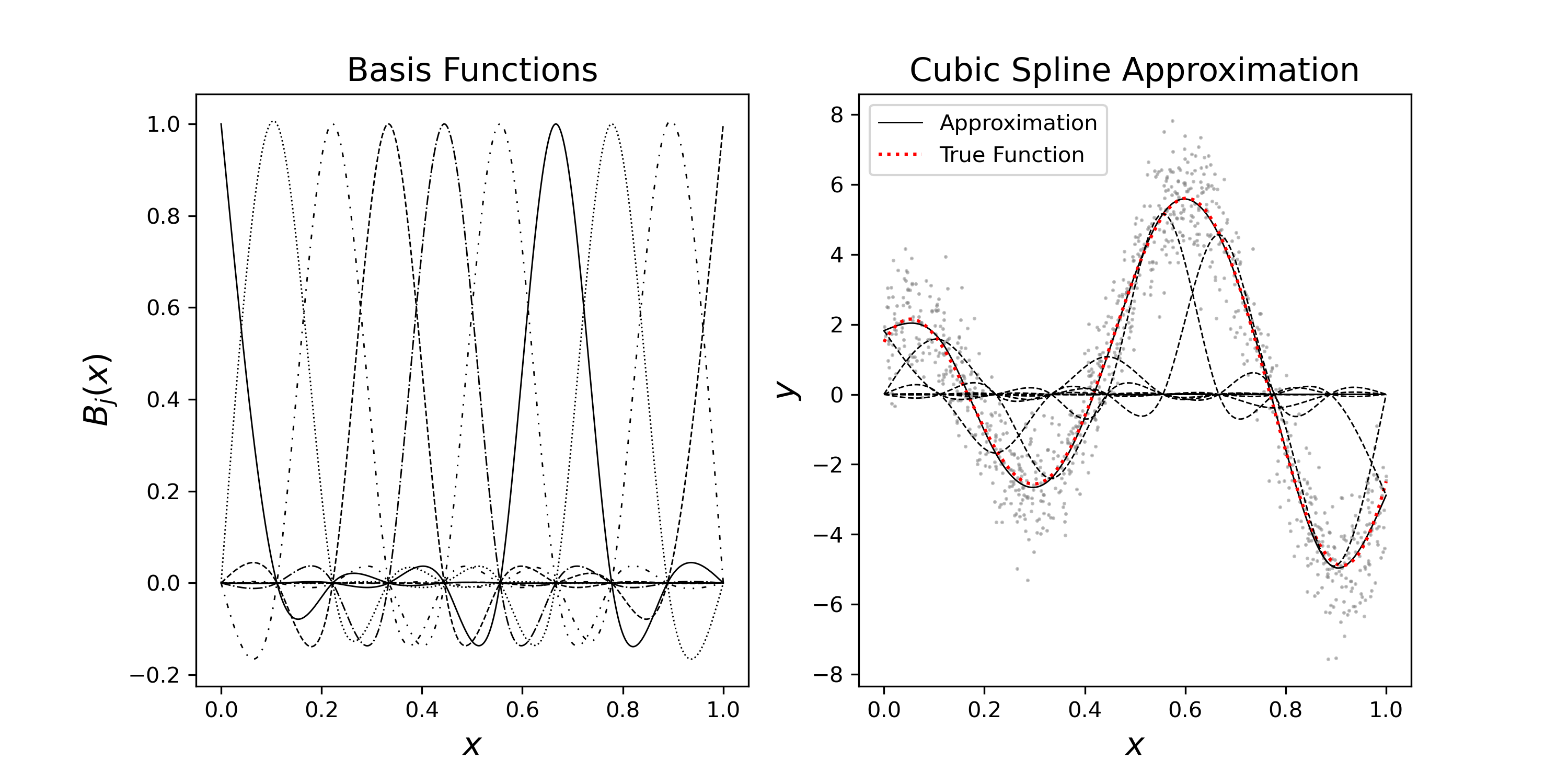}
\caption{Visualization of the cubic regression splines. The left hand side displays the basis functions and the right hand side shows how the function approximation is done by scaling the bases appropriately }
    \label{fig:spline_approximation}
\end{figure} 

For the construction of the smoothers, $f_i(x_i)$,  a basis expansion with nonlinear bases is applied and the final function is represented as  their linear combination:
\begin{equation} \label{eq:lincom}
f(x_i) = \sum_j^k B_j(x_i)\beta_j.
\end{equation}
The location of the basis functions, $B_j$, is parametrized by an ordered set of knots $\kappa_1< \kappa_2 < ... < \kappa_k$, where the positioning is traditionally chosen to be either uniformly distributed over the input domain or located at the positions of the quantiles. The later one allows the model to be more flexible in regions with more data and is more restrictive in sparse areas. 

To control the smoothers ``wigglyness'' and hence overfitting their squared second derivative can be penalized. This allows a trade-off between fit to the data and good generalization. However, due to their smoothness the splines are by design already robust against overfitting and since they are taking their local neighborhood into account, they also yield good results on respectively small datasets \cite{wood_generalized_2017}.
Parameter estimation is hence done by penalized likelihood maximization, which aims to solve the following optimization problem:
\begin{equation} \label{eq:optimization}
    \underset{\theta}{\text{arg min}} - \frac{1}{n} l(\theta,x,y) + \sum_j^k \lambda_j \int f''(x)^2dx,
\end{equation}
with $\theta=\left [\beta_1, \dots ,\beta_k \right]$
with penalized iterative reweighted least squares (PIRLS).
$l(\theta,x,y)$ defines the log-likelihood and $\lambda_j$ determines the impact of the penalization of the ``wigglyness''. Further hyperparameters are the positions of the knots $\kappa$, as well as the number of basis functions. Theoretically a higher number of bases leads to more flexible functions. In practice, the dependence on the number of bases is only moderately big, as it is compensated by the penalization \cite{wood_generalized_2017}.

\begin{figure*}
         \centering
         \includegraphics[width=\textwidth]{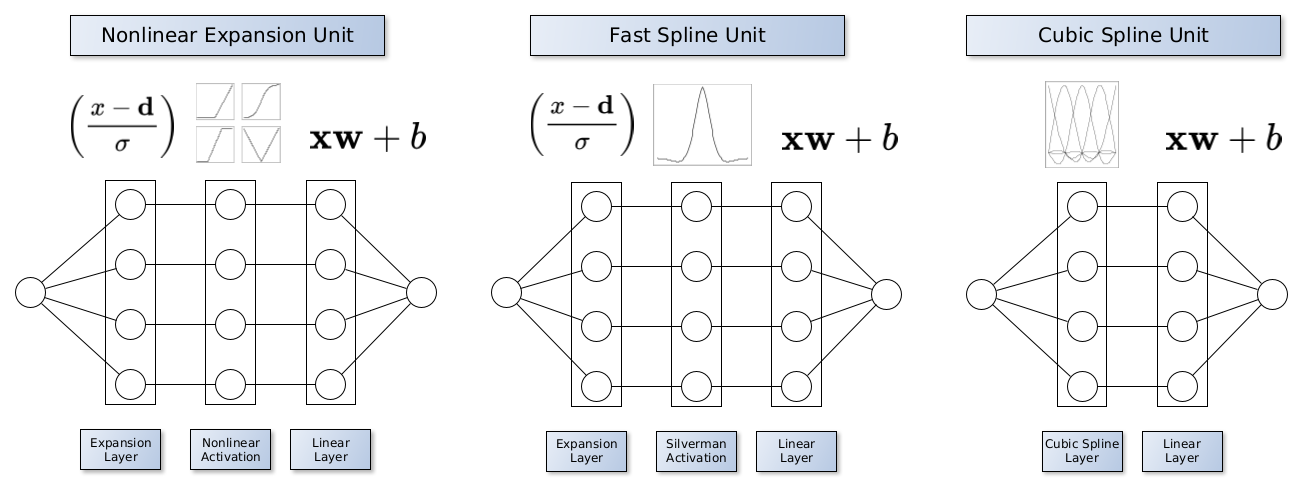}
         \caption{The figure sketches how the different types of approximations could be implemented as neural networks.}
         \label{fig:generalized_layers}
\end{figure*}

\subsection{Basis Functions and Activation Functions} 
To model explainable terms in neural networks with smooth functions, we propose spline units according to equation \ref{eq:lincom}. Hence, a unit consists of a basis expansion layer in connection with an ordinary linear layer to form the weighted sum. For simplicity, we introduce the model framework based on the cubic spline unit, where the basis expansion layer is constructed around the $\mathbf{G}$ matrix similarly as the cubic expansion in \cite{wood_generalized_2017}:

\begin{equation}
    \mathbf G =
    \begin{pmatrix}
    \mathbf 0 \\
    \mathbf{\mathbf{B}^{-1}}\mathbf{D} \\
    \mathbf{0}
    \end{pmatrix},
\end{equation}

with  $h_j=\kappa_{j+1}-\kappa_j$, 
\begin{align*}
\mathbf{B} =
\begin{psmallmatrix} 
\frac{1}{2}(h_1+h_2) & \frac{1}{6}h_2 & 0 & \dots & 0 \\
\frac{1}{6}h_2 & \frac{1}{3}(h_2+h_3) & \frac{1}{6}h_3 & \dots & 0 \\
0 & \ddots & \ddots &  & \vdots\\
\vdots &  &  & \ddots & \frac{1}{6}h_{k-2} \\
0 & \dots & 0 & \frac{1}{6}h_{k-2} & \frac{1}{3}(h_{k-1}+h_k-1)
\end{psmallmatrix}
\end{align*}
and\\
\begin{align*}
\mathbf{D}=
\begin{psmallmatrix}
 \frac{1}{h_2} & -\frac{1}{h_1}-\frac{1}{h_2} & \frac{1}{h_2} & 0 & \dots & 0 \\
 0 & \frac{1}{h_2} & -\frac{1}{h_2}-\frac{1}{h_3} & \frac{1}{h_3} &  & \vdots \\
 \vdots &  & \ddots & \ddots & \ddots & \\
 0 & \dots & 0 & \frac{1}{h_{k-2}} & -\frac{1}{h_{k-2}}-\frac{1}{h_{k-1}} & \frac{1}{h_{k-1}}
\end{psmallmatrix}.
\end{align*}
Based on that, the expansion is a function $B: \mathbb{R} \rightarrow \mathbb{R}^k$ with the explicit formulation:
\begin{equation} \label{cubicconstruction}
\begin{split}
    B(x) = (c^-_j \mathbf{G}_{j,\bullet} + a^-_je_j + c^+_j\mathbf{G}_{j+1,\bullet}+ a_j^+e_{j+1})
\end{split}
\end{equation} 
\begin{align*} 
c^-_j = \frac{(\kappa_{j+1}-x)^3}{h_j} - \frac{h_j(\kappa_{j+1}-x)}{6} & & a_j^- =\frac{(k_{j+1}-x)}{h_j}\\
    c^+_j = \frac{(x-\kappa_{j})^3}{h_j} - \frac{h_j(x-\kappa_{j})}{6} & & a_j^+ =\frac{(x-k_{j})}{h_j}\\
    h_j = \kappa_{j+1}-\kappa_{j} & & j \text{ s.t. } \kappa_j < x < \kappa_{j+1}
\end{align*}
for values $ x \in [\kappa_1,\kappa_k]$, whereby the extrapolation is handled separately. Furthermore, $e_j$ represents the $j$-th standard basis.
Eventually $B(x)$ is fully defined by the knot positions and thus it can be implemented straightforward as a parametrized layer with fixed knots.
In case of the penalized splines the layer is further equipped with the smoothing matrix $\mathbf{S}=\mathbf{D}^T\mathbf{B}^{-1}\mathbf{D}$ which can be used to determine the penalty.

\subsubsection{Fast Spline Units}
The cubic splines represent a coherent way to achieve nonlinearity in a model and come with convenient properties such as a guaranteed smoothness of the resulting function, a good fit and implicit penalization. In context of deep learning it could be viewed as a special activation layer, but in contrast to the established activations such as ReLU or Sigmoid, the gradient calculation is more expensive. 
This section generalizes the idea of basis expansion to ordinary activation functions and hence shows the generalizability of the presented approach. Fast approximations of the cubic spline layer based on the Silverman kernel \cite{silverman_spline_1984} are derived.
Since polynomial splines can be represented as a truncated power series of degree $d$, (\ref{eq:lincom}) can be expressed as a polynomial part and a truncated polynomial part, with $\kappa_1 < \dots < \kappa_k$ representing a sequence of knots \cite{fahrmeir_regression_2013}:
\begin{equation}
\begin{split}
    f(x)= \beta_0 + \beta_1 x + \dots + \beta_{d}x^d + \beta_{d+1}(x-\kappa_2)_+^d + \dots + \\  \beta_{k+d-2}(x-\kappa_{k-1})_+^d
\end{split}
\end{equation}
with 
\begin{equation}
    (x-\kappa)_+^d = 
\begin{cases}
    (x-\kappa)^d,& \text{if } x\geq \kappa\\
    0,              & \text{otherwise}.
\end{cases}
\end{equation}
Especially for the truncated part the idea is the same as for the cubic splines in a way that the input domain is populated with shifted versions of the basis function. 
\begin{figure}
\includegraphics[width=.5\textwidth]{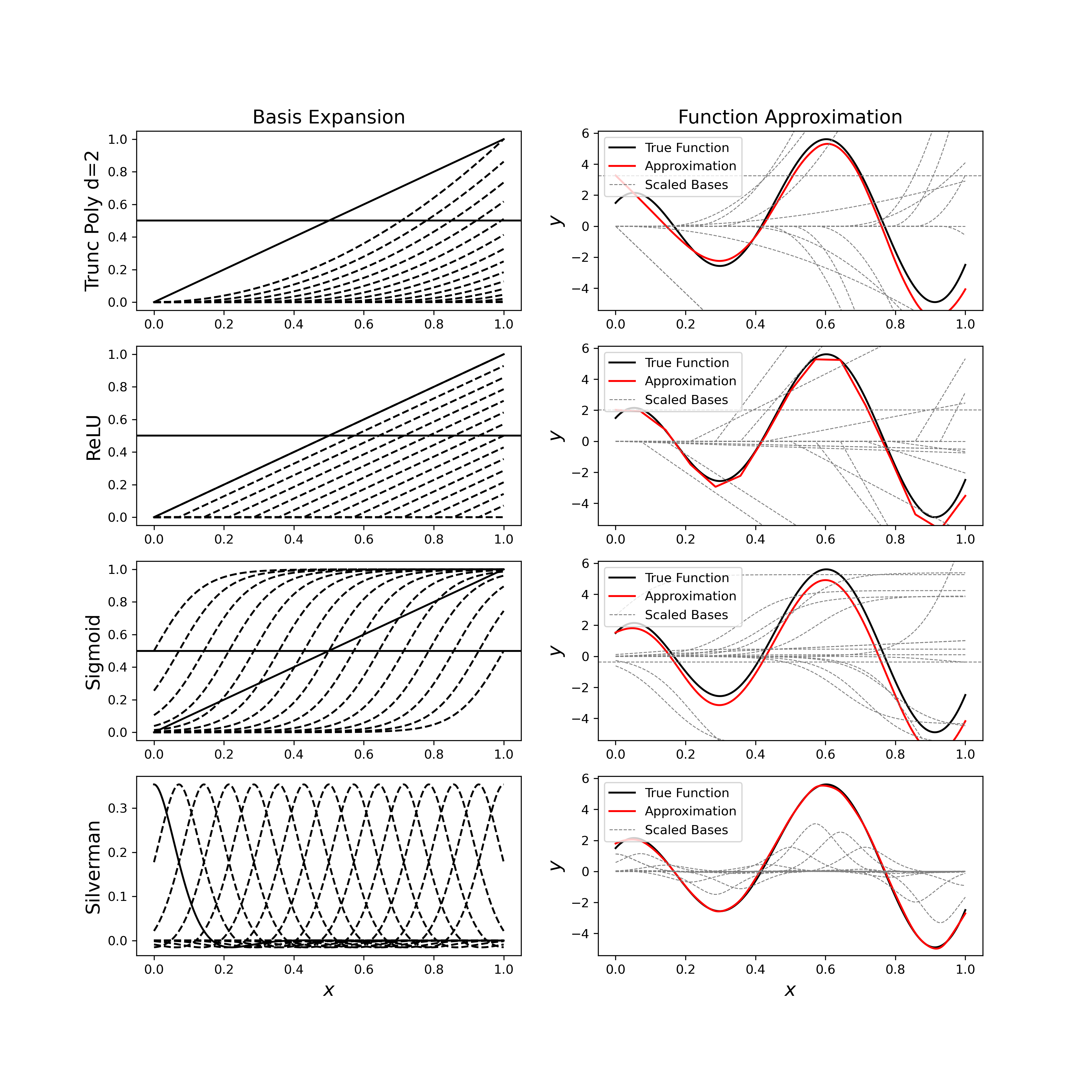}
\caption{Basis expansion based on different types of activation functions along with their approximations of an exemplary function.}
\label{fig:comparison_basis_expansion}
\end{figure} 
The essential difference is that truncated polynomials are used instead of the bell-shaped cubic bases. The approximation is then again performed by searching for an appropriate scaling, such that the bases sum up roughly to the function of interest  (see figure \ref{fig:comparison_basis_expansion}).
For $d=1$ the truncated polynomials are actually equivalent to shifted ReLU functions and thus the exact same behaviour could be accomplished by shifting the data rather than the function and passing it through an ordinary ReLU. From that point it is easy to see how the idea of basis expansion can be generalized to various types of activations, which is also visualized in figure \ref{fig:comparison_basis_expansion}.
To get closer to the cubic splines the Silverman kernel is used as an activation function.
\begin{equation}
    K(x) = \frac{1}{2}e^{\frac{-|x|}{\sqrt{2}}}\text{sin} \left(\frac{|x|}{\sqrt{2}} + \frac{\pi}{4} \right)
\end{equation}
It is considered to be the kernel method that is equivalent to cubic smoothing splines \cite{silverman_spline_1984} and thus has the same characteristic bell shape as the cubic spline bases. One difference is that the width of the basis is not determined by some knot positions and has to be handled by adjusting the bandwidth of the kernel, or respectively scaling the data. More precisely, the data is passed through the kernel as $K(\frac{x-d}{\sigma})$ with $d$ as the displacement and $\sigma$ as the scaling. Other properties that are lost by approximating the cubic spline unit with Silverman kernels, are the implicit control of the smoothness and the linear extrapolation, but on the other hand the calculation of the gradients becomes easier and furthermore the position of each kernel as well as its bandwidth can easily be set to be learnable parameters.

\subsubsection{Learnable Knots}
Modeling jagged shape functions and accurately representing jumps is an advantage of deep neural networks because of their flexibility. Multi-layer perceptrons can accurately represent jumps using an appropriate number of neurons and appropriate activation functions. In contrast, \cite{caruana2015intelligible} found that GAMs can overregulate and do not accurately represent details in data. Flexible choice of node locations, however, in addition to an appropriate number of knots, can easily overcome this problem \cite{mohanty2021adaptive, galvez2011efficient}.
Since the expansion can be broken down to a couple of simple arithmetic operations and matrix products in respect to $\kappa$, backpropagation can be used to make the knot positions trainable parameters. The only part that requires extra care is the assumption that the knots are an ordered sequence $\kappa_1 < ... < \kappa_k$. Thus, before each forward pass the knot parameters have to be sorted appropriately and the $\mathbf{F}$ matrix has to be recalculated accordingly. 
In equation \ref{cubicconstruction} it can be seen that the construction of $\mathbf{F}$ involves the division by $h_j=\kappa_{j+1}-\kappa_j$, which can lead to numerical issues when two neighbouring knots are getting too close and can be circumvented with a knot distance penalty $g(\kappa) = \sum_i^{k-1} \frac{1}{h_j}$.

\begin{figure}
\includegraphics[width=.5\textwidth]{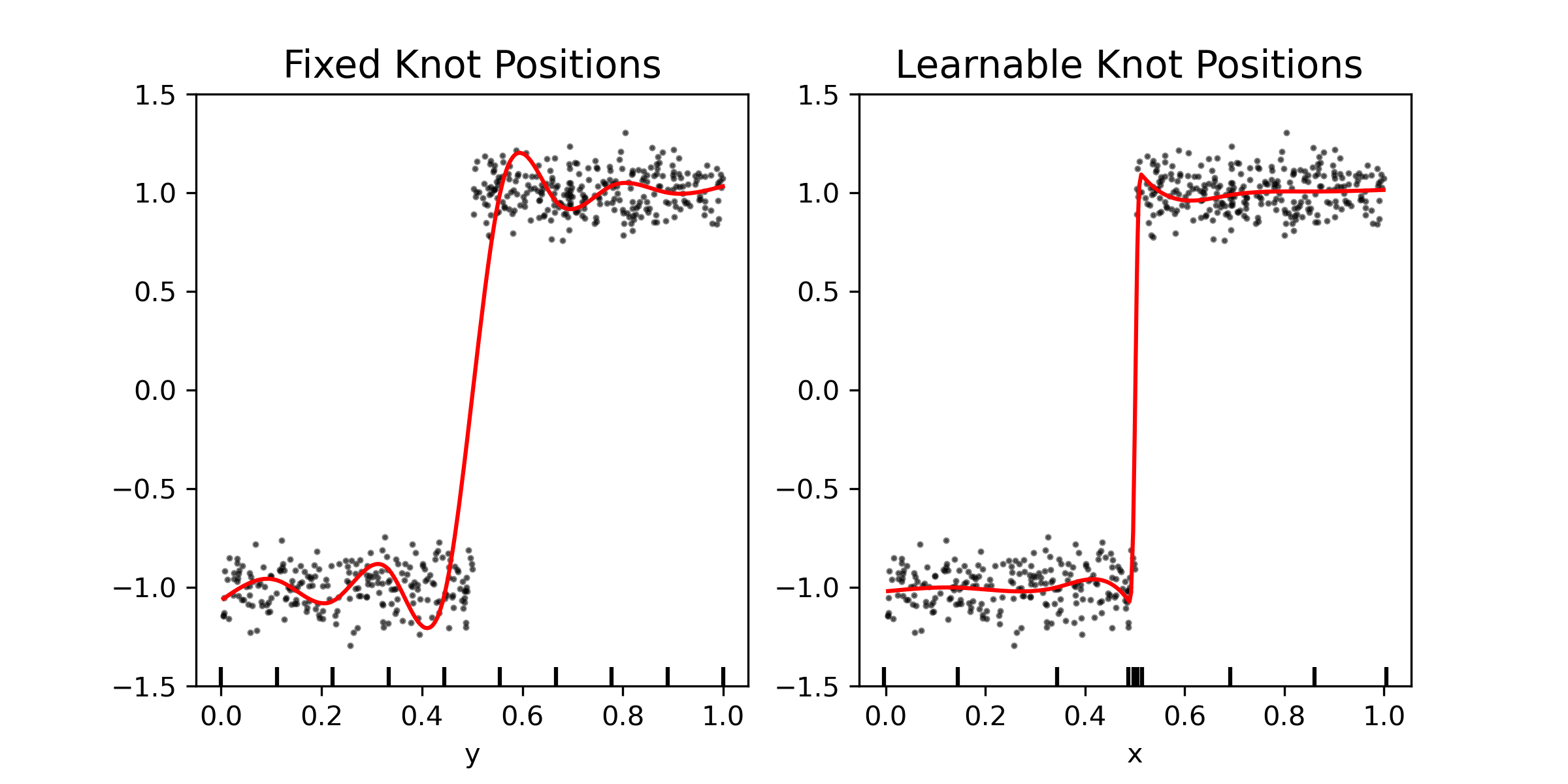}
\caption{Turning the knot positions into learnable model parameters, enables the SNAMs to fit jagged functions. The black marks on the x-axis indicate the knot positions after training.}
\label{fig:LearnableKnots}
\end{figure} 

\subsection{Identifiability}
Without further restrictions, the identifiability is not fulfilled for GAMs and hence not fulfilled for SNAMs. One could easily shift an additive constant from one component $f_i(x)$ to another $f_j(x)$ without affecting the predictive behaviour of the model at all. To prevent this, additional constraints are imposed which ensure that \begin{equation}
\sum_i^n f(x_1)= \dots = \sum_i^n f(x_k) =0
\end{equation} 
and all the additive constants are collected in a global intercept. Those constraints do not affect the capabilities of the model, but ensure the desired uniqueness and full interpretability of its components.
One way of implementing such a constraint is by subtracting the column means of the basis expansion. To make the presented units identifiable an intermediate layer between the expansion and the summation is implemented which tracks the running column means of the expansion and subtracts it during each forward pass.

\section{Enhancing interpretability with SNAMs}
One major advantage of SNAMs is that they offer a toolbox for improving intelligibility of the model while retaining the high-accuracy of state-of-the-art DL models. While NAMs turn black-box DL models into glass-box models mainly by visualization, SNAMs can be considered as sand-box models. This highlights the fact that SNAMs can be used for a post-hoc analysis widening the understanding of the underlying model's behaviour and therefore the data-generating process. These improvements can be connected to different properties described in this section.
%enhanced intelligibility further  
%benefit, advance, state-of-the-art, high-accuracy 
%reduce high variance, uncertainty estimates, improvement
%expressivity
\subsection{Low dimensional parameter space}
Standard DL models and also recently proposed NAMs are notoriously overparameterized. Owning their flexibility to the high dimensionality, they often consist of hundreds of thousands or even millions of parameters. On the contrary SNAMs considerably reduce the number of parameters to the amount of a few hundred or so.\\

The stability of the fitting process is largely improved with less multiple and more pronounced minima of the loss landscape, leading to more reliable parameter estimates. 
Furthermore the parameters are intelligible as the weights scale local basis function, see formulation (\ref{eq:lincom}), while the biases define the knot positions. This allows the local behaviour of each single feature to be investigated and for instance formally specified by a derivative. This well-behaved parameterization is precisely the reason for the openness of SNAMs to methods from the statistical community such as calculating confidence bands, hypothesis testing or using AIC / BIC based model selection.\\

SNAMs maintain the well calibrated estimates between over- and underfitting for low to moderate sized datasets. This supersedes the need to distinguish between ``too small'' and ``large enough'' data DL models are exposed to.

\subsection{Understanding spatial effects}
Although limiting the amount of parameters SNAMs can model interactions in high-dimensional feature spaces if need be. This is especially useful for spatial data.  While in the NAM framework for the California Housing dataset longitude and latitude are included separately \citep{agarwal2021neural} both can be included in an interaction term by SNAMs.\\ 
Two (or more) dimensional smoothers can be modelled as tensor product splines by evaluating two univariate smoothers along each dimension and estimating a coefficient for the combinations of their Cartesian product
\begin{equation}\label{tpspline}
    f_{ij}(x_i,x_j)=\sum_{l,r}^{p,q} B_l(x_i)B_r(x_j)\beta_{l,r}.
\end{equation}

The learned shape function $f_{ij}(x_i,x_j)$ can be visualized as a 2d heat map for the California Housing data, see figure (\ref{tpspline}). The effect size is given by the color with yellow indicting a strong positive effect and blue indicting a strong negative effect. This gives human interpretable effect strengths and provides a sense of confidence with regard to the $p\times q$ learned parameters.
 
\begin{figure}
    \centering
        \includegraphics[width=0.4\textwidth]{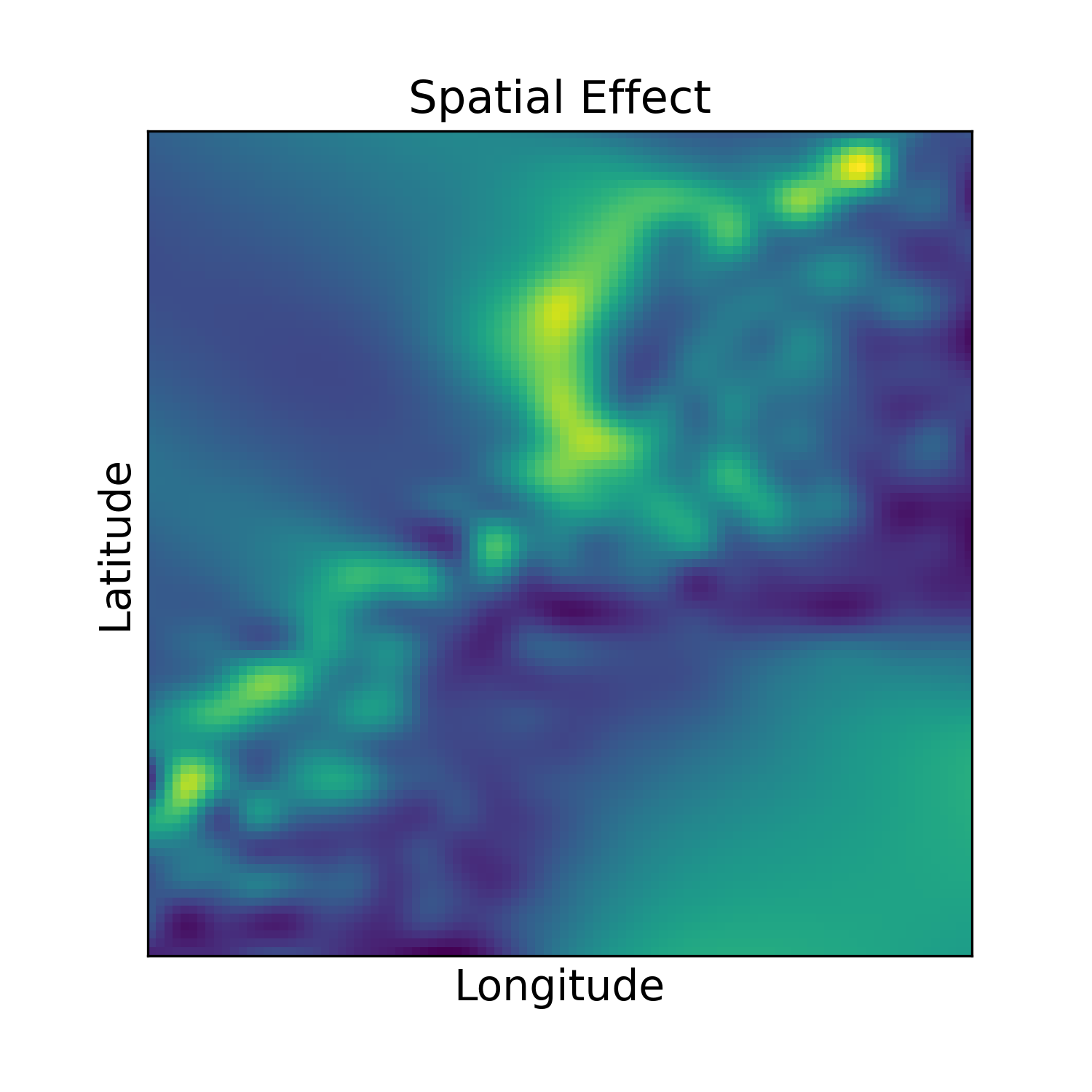}
        \caption{The results of the SplineNAM on the CA Housing dataset displaying a heatmap of the spatial effect.}
        \label{fig:spatial_results_ca}
\end{figure}

\subsection{Intelligibility beyond visualization}
Intelligibility goes far beyond visualization. While with NAMs effects of single features can be visualized, no further statements on the underlying data generating process can be made. However with SNAMs further intelligibility is available. This is due to the low dimensional parameter space giving meaning to the parameters. This allows for the quantification of the parameter uncertainty. As a result meaningful confidence bands can be calculated and questions regarding the presence or absence of feature effects can be answered withing the hypothesis testing framework. The key quantity to asses the uncertainty is the Fisher information. \\

\textbf{Fisher Information}: The Fisher information matrix is one of the core concepts in the maximum-likelihood framework. It is defined as the covariance of the gradient of the log-likelihood. Hence the Fisher information is defined as

\begin{equation}
    \mathbf{F}(\theta) = E \left(\frac{\delta l(\theta,x,y)}{\delta\theta}\frac{\delta l(\theta,x,y)}{\delta\theta}^T \right).
\end{equation}
Furthermore under certain regularity conditions the Fisher information can be expressed as second partial derivatives the log-likelihood, i.e. the Hessian \cite{ly_tutorial_2017}
\begin{equation}
    \mathbf{H(\theta)} = -E \left(\frac{\delta^2l(\theta,x,y)}{\delta\theta\delta\theta^T} \right).
\end{equation}

\textbf{Confidence Bands}: In context of parameter estimation the Fisher information is used together with the asymptotic distribution of 
\begin{equation}
    \betahat \overset{asy}{\sim} N \left(\beta, \mathbf{F}^{-1}(\beta)\right)
\end{equation}
 to asses the uncertainty of the parameters $\hat{\beta}$. In the classic GLMs $\alpha$-level confidence intervals are in general constructed as 
 \begin{equation}
 \left[\betahat_i \pm z_{1-\frac{\alpha}{2}}\sqrt{\FIMinv_{ii}}\right]
 \end{equation} 
 \cite{hastie_elements_2009} with $z_{1-\frac{\alpha}{2}}$ as the $(1-\frac{\alpha}{2})$ quantile of the standard normal distribution. Even though theoretically the confidence bands for the GAMs could be derived in the same pointwise manner for each of the coefficients, in practice this leads to poor coverage results since it neglects the structure between basis functions and thus, the construction of Bayesian credibility bands is preferred \cite{wood_generalized_2017}.
Hence, new coefficients are sampled from the posterior
\begin{equation}\label{eq:sample_ci}
    \beta^* \sim N\left(\betahat, (\FIM+ \lambda \mathbf{S})^{-1}\right)
\end{equation}
and the bands are represented by the quantiles of the resulting functions 
\begin{equation}
f_i^*(x)= \sum_j^k B_{j}(x)\beta^*_j
\end{equation}
with smoothing matrix $\mathbf{S}$ as in (\ref{sec:cubic_spline_unit}).
From a deep learning perspective the Fisher information is especially interesting in context of optimization, as it can capture the local geometry of the loss landscape \cite{pmlr-v89-karakida19a} and thus helps to improve the gradient steps. 
Since the true Fisher information is not always easy to derive, but the gradients are calculated anyways during the optimization, the so-called empirical Fisher information is used instead to estimate $\FIM$ as an averaged outer product of the (stochastic) gradients \cite{martens_new_2014, soen_variance_2021}

\begin{equation}\label{empirical_fisher}
    \mathbf{\Tilde{F}}(\theta) = \frac{1}{N}\sum_i^N \left[ \Delta_\theta l(\theta,x,y)\Delta_\theta l(\theta,x,y)^T \right]
\end{equation} 

While the empirical Fisher is easy to derive, in practice it is barely used directly in deep learning since the storage capacity grows quadratic with the number of parameters. This makes it infeasible for ordinary neural networks and instead diagonal- or factorized approximations are used to cope with the complexity, but those approaches are still under active research \cite{martens_optimizing_2015, frantar_m-fac_2021}. Here, the extreme parameter sparsity of the SplineNAMs pays off since $\Tilde{\mathbf{F}}$ is actually feasible for reasonable settings.\footnote{\cite{kunstner_limitations_2019} provides a critical discussion of the use of the empirical Fisher and even though he admits that it can improve the optimization he also points out, that it is only a consistent estimate for the Fisher information close to the minimum, with a sufficient amount of data and for a correctly specified model.
Especially the first two conditions justify the use of the empirical Fisher to construct confidence intervals for the SplineNAMs as it is only evaluated after the fitting procedure and respectively few parameters are involved.}

\section{Benchmarking the Accuracy of SNAMs}\label{sec:real}
\begin{table*}
\caption{\textbf{Results of the benchmark study}: We orientate on the benchmark study conducted in \citet{agarwal_neural_2021} and compare SNAMs to state-of-the-art neural as well as non-neural approaches. We report the means and standard deviations of 5-fold cross validation on multiple datasets. We report the results on widely used and publicly available datasets. A detailed description on the datasets as well as the preprocessing steps can be found in the appendix \ref{app:data}. }
\begin{tabular}{|l|l||c|c||c|c|}
\hline
\textbf{Model} & \textbf{No. Params} & \textbf{CA Housing} (RMSE)$\downarrow$  & \textbf{Insurance} (RMSE) $\downarrow$ & \textbf{Credit} (AUC) $\uparrow$ & \textbf{FICO} (AUC) $\uparrow$ \\
\hline
\hline
MLP         & $> 90,000$ & 0.433 $\pm$ (0.009)  & 0.425 $\pm$ (0.031) & 0.942 $\pm$ (0.012) & 0.795  $\pm$ (0.006)  \\ \hline
XGBoost     & - & 0.403 $\pm$ (0.008)  & 0.427 $\pm$ (0.025) & 0.961 $\pm$ (0.022) & 0.714  $\pm$ (0.008)   \\ \hline
EBM         & - & 0.441 $\pm$ (0.011)  & 0.384 $\pm$ (0.025) & 0.954 $\pm$ (0.018) & 0.727  $\pm$ (0.011)  \\ \hline
NAM         & $> 500,000$ & 0.483 $\pm$ (0.011)  & 0.497 $\pm$ (0.028) & 0.904 $\pm$ (0.012) & 0.799  $\pm$ (0.008)  \\ \hline
NAM(ExU)    & $> 500,000$ & 0.510 $\pm$ (0.011)  & 0.499 $\pm$ (0.029) & 0.913 $\pm$ (0.013) & 0.790  $\pm$ (0.008)  \\ \hline
\hline
SNAM  & $<2.400$  & 0.396 $\pm$ (0.009) &  0.487 $\pm$ (0.039) &  0.948 $\pm$ (0.011) &   0.782 $\pm$ (0.011)\\
\hline
FAST SNAM & $<9,600$  & 0.390 $\pm$ (0.016) &   0.482 $\pm$ (0.049) &  0.968 $\pm$ (0.005) &   0.788 $\pm$ (0.008)\\\hline
\end{tabular}

\label{table:benchmark}
\end{table*}

In this section we asses the predictive performance of SNAMs in comparison to several state-of-the-art models including neural as well as non-neural approaches. The following four datasets were chosen for this purpose to reflect real-world uses cases where model interpretability is a key factor:
\begin{itemize}
    \item \textbf{California Housing:} A widely used regression dataset for predicting home prices in California from the U.S. census in 1990.
    \item \textbf{Insurance:} A regression dataset for predicting billed medical expenses containing 1338 observations.
    \item \textbf{Credit:} A fraud detection dataset with a binary response and heavily unbalanced class distributions.
    \item \textbf{FICO:} A classification dataset for modelling credit score predictions.
\end{itemize}
A detailed description of the datasets and the contained features can be found in the Appendix. The data pre-processing was adapted from the benchmark study of Gorishniy \cite{gorishniy_revisiting_2021}, who evaluated various DNNs on tabular data. In particular a quantile scaling was applied to all features and the target variable was centered to zero mean and standardized to unit variance. For The regression problems (California Housing, Insurance) the averaged RMSE over a 5-fold cross validation is reported in table \ref{table:benchmark} and for the classification problems (FICO, Credit) the AUC is taken respectively.

We used the following benchmarks to asses the performance of the presented method:
\begin{itemize}
    \item \textbf{Multilayer Perceptron (MLP)}: Unrestricted fully connected deep neural network. The loss function is either defined as the mean squared error (regression) or binary cross entropy (logistic regression).
    \item \textbf{Gradient Boosted Trees (XGBoost)}: Decision tree based gradient boosting. We use the implementation provided by \cite{Chen:2016:XST:2939672.2939785}.
    \item \textbf{Explainable Boosting Machines (EBMs)}: State-of-the-art Generalized Additive Models leveraging shallow boosted trees \citep{lou2013accurate}.
    \item \textbf{Neural Additive Models (NAMs)}: Linear combination of deep neural networks that allow visual interpretability. Each feature is modeled by a different subnetwork. We implement two NAM versions, one leveraging \textit{exp-centered} (ExU) hidden units \citep{agarwal_neural_2021} and one using ReLU activation functions throughout the hidden layers.
\end{itemize}

%\begin{table*}
%\begin{tabular}{|l|c|c|c|}
%\hline
%\textbf{Model} & \textbf{No. Parameters} & \textbf{CA Housing} (RMSE) & \textbf{Dataset II...} \\
%\hline
%\hline
%TabNet   & - &0.510  $\pm$ (7.6e-3)  & {}      \\ \hline
%SNN      & - &0.493  $\pm$ (4.6e-3)  & {}     \\ \hline
%AutoInt  & - &0.474  $\pm$ (3.3e-3)  & {}     \\ \hline
%GrowNet  & - &0.487  $\pm$ (7.1e-3)  & {}     \\ \hline
%MLP      & - &0.499  $\pm$ (2.9e-3)  & {}     \\ \hline
%DCN2     & - &0.484  $\pm$ (2.4e-3)  & {}     \\ \hline
%NODE     & - &0.464  $\pm$ (1.5e-3)  & {}     \\ \hline
%ResNet   & - &0.486  $\pm$ (2.9e-3)  & {}     \\ \hline
%CatBoost & - &0.430  $\pm$ (7.4e-4)  & {}     \\ \hline
%XGBoost  & - &0.433  $\pm$ (1.6e-3)  & {}     \\ \hline
%\hline
%SplineNAM (ours) & -  & 0.434 $\pm$ (7.4e-3) & {} \\ \hline
%\end{tabular}
%\end{table*}

The results show that the SNAMs are all throughout able to reach at least the same performance range as traditional NAMs and even outperform them on California Housing, FICO and Insurance. The insurance dataset has respectively few observations(N=X). In this case the improvement of the SNAMs might be attributed to their parameter sparsity and consequently a enhanced small-sample behaviour. 
However, in particular noteworthy is the substatial performance gain that the SNAMs achieved on the California Housing dataset. This is likely due to the SNAM's ability to jointly process the geographical coordinates as a spatial effect and highlight the importance of integrating structured data appropriately into DNNs.

\subsection{Interpretability}
\begin{figure}[h]
    \centering
        \includegraphics[width=0.5\textwidth]{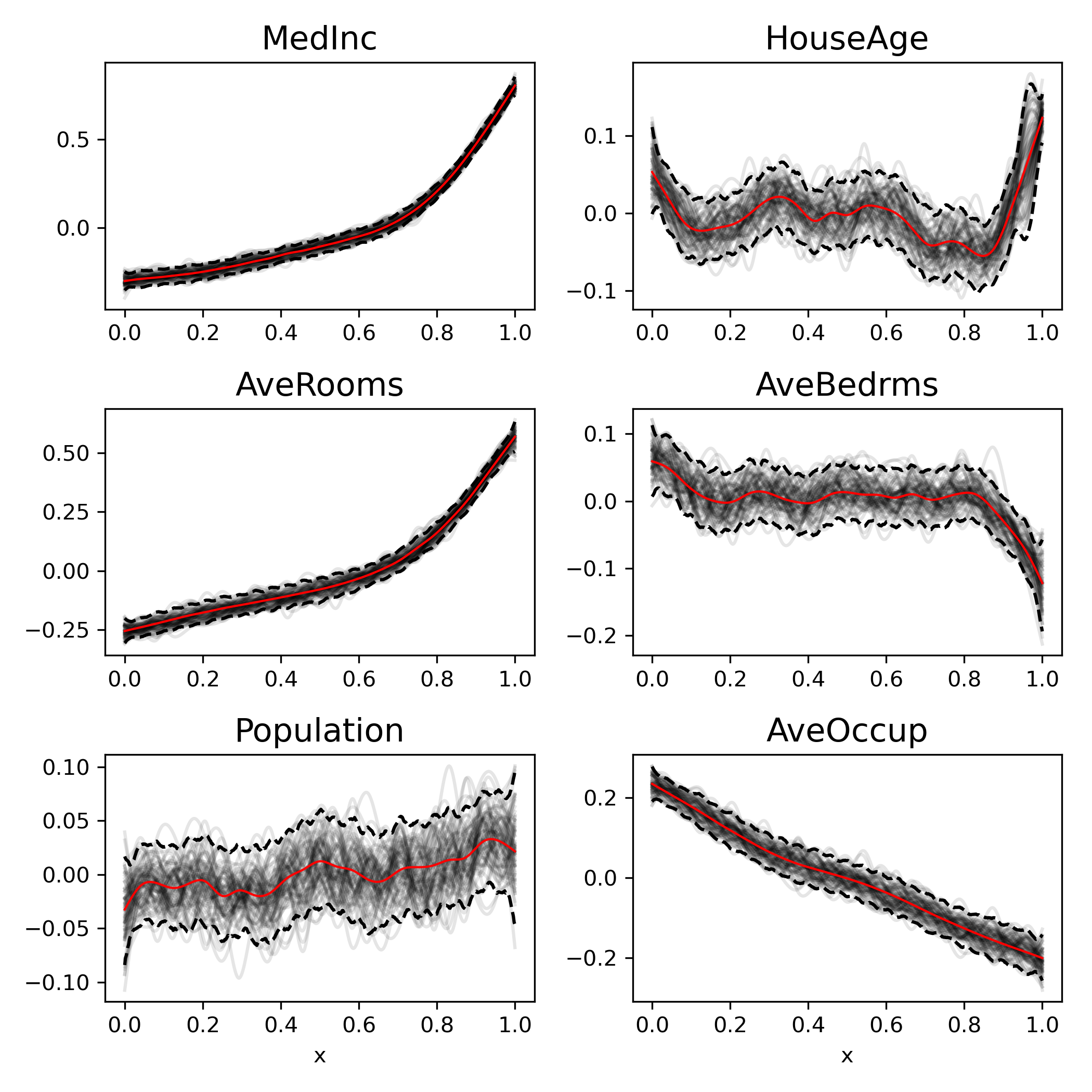}
        \caption{The results of the SNAM on the CA Housing dataset showing the estimated nonlinear effects of the features with Bayesian confidence intervals.}
        \label{fig:feature_results_ca}
\end{figure}

While achieving remarkable results on the benchmark datasets, SNAMs remain to be fully explainable glass-box-models and not only allow to investigate the exact decision making process, but  yield the identifiable effect of each feature as shown in figure \ref{fig:feature_results_ca}. The spatial features result in a 2D geographical heatmap, see figure \ref{fig:spatial_results_ca}. That indicates which areas have a respectively positive or negative effect on the house price. Usually, those estimations could be interpreted directly, but due to the quantile scaling of the features, this interpretation has slight restrictions in terms of the scale of the x-axis. Nevertheless, valuable conclusions can be derived such as that the median income, as well as the average rooms have an increasing positive effect, whereas for the average number of household members it is in reverse. The house age indicates that in certain timespans more valuable houses were built and the average number of bedrooms seems to not have a big effect, except for large numbers, which decreases the house value.

%\section{Stable fitting procedures for interpretability}

\section{Related Work}
Achieving interpretability in deep neural networks is inherently difficult. While several approaches have successfully created interpretability at the sample level \citep{shapley1953quota, sundararajan2020many, baehrens2010explain, ribeiro_why_2016}, the interpretability of classical statistical generalized additive models have yet to be matched in a neural framework. 
The idea of transferring the advantages of generalized additive models to neural approaches  was already introduced by \citet{potts1999generalized}. However, while being remarkably parameter sparse, the framework did not use backpropagation and was therefore ultimately unsuccessful in implementing the GAM structure into a neural framework. \citet{de2007generalized} translate GAMs into a neural framework. More recently, \citet{agarwal_neural_2021} introduced NAMs, a more flexible approach than the GANNs introduced by \citet{de2007generalized} and exploiting the advances made in the field of deep learning. Similar to SNAMs, NAMs learn the non-linear additive functions with seperate networks for each feature. Using large numbers of hidden units for each feature and exploiting flexible exponentially centered activation functions, they successfully integrate the predictive power of neural networks into an additive framework. However, while the linear combination of neural sub-networks allows for good visual interpretation of the results, any interpretability beyond plotting the model predictions on a feature level is lost in their black-box sub-networks. Additionally, the predictive power comes at the cost of an enormous number of parameters and a huge hyperparameter space for optimization.
Building upon NAMs, several extensions have been proposed. \citet{chang2021node} introduced NODE-GAM, extending the works of \citet{lou_intelligible_2012} a differentiable oblivious decision trees based model designed for high-risk domains. \citet{yang2021gami} expand NAMs to adjust for pairwise interaction effects. \citet{rugamer_semi-structured_2020} extends statistical structural additive models by neural networks constructing an orthogonalization cell. \citet{thielmann2023neural} extend the framework to account for arbitrarily many distributional parameters.

\section{Conclusion}% and Future Work}
In this work we present Structural Neural Additive Models as an intelligible extension of the recently proposed Neural Additive Models. We demonstrate the integration of basis expansion layers into neural networks as a way to flexibly model nonlinear functions with extremely few parameters. Furthermore, we derive computationally efficient neural splines based on the Silverman kernel. In a benchmark study we show that this approach is able to reach the same predictive performance as other state-of-the-art models, while remaining inherently more interpretable.  In conclusion, Structural Neural Additive Models (SNAMs) offer a promising architecture for interpretable neural networks, combining classical statistical methods with the predictive power of neural architectures.

\bibliographystyle{icml2023}
\bibliography{bib}

%%%%%%%%%%%%%%%%%%%%%%%%%%%%%%%%%%%%%%%%%%%%%%%%%%%%%%%%%%%%
%\section*{Acknowledgements}
%Funding by the Deutsche Forschungsgemeinschaft (DFG, German Research Foundation) within project 450330162 is gratefully acknowledged.

\appendix
\onecolumn

\section{Benchmarking}

\subsection{Data sets \& Preprocessing}\label{app:data}
\paragraph{California Housing}
The California Housing (CA) dataset \cite{pace_sparse_1997} contains the house prices for areas in California from the U.S. census in 1990. The data was obtained from sklearn \cite{pedregosa_scikit-learn_2011} and  besides the logarithmic median house price of the blockwise areas as the target variable it contains the eight predictors from the following list.
Of particular interest is, that the CA dataset contains the longitude and latitude as spatial features. Considering that those features are more meaningful when they are interpreted mutually, they can serve as a good example for a usecase of tensor product splines in SplineNAMs.  Besides that, the remaining features are modelled univariately with cubic spline units and for the sake of simplicity a sophisticated hyperparameter tuning is omitted. Instead the number of bases and feature-wise penalties are chosen heuristically by trying out a small number of configurations on a single training set.

\begin{itemize}
    \item \textbf{Targets:} Median House Price
    \item \textbf{Features:}
    \begin{enumerate}
        \item \textit{MedInc}:    Median income of the area                             
        \item \textit{HouseAg}e:  Median house age of the area                          
        \item \textit{AveRooms}:  Area-wise average number of rooms per household       
        \item \textit{AveBedrms}: Area-wise average number of bedrooms per household    
        \item \textit{Population}:Total number of people living in the area             
        \item \textit{AveOccup}:  Area-wise average number of people within a household 
        \item \textit{Latitude}:  Latitude of the center of the area                    
       \item \textit{Longitude}: Longitude of the center of the area                     

    \end{enumerate}
\end{itemize}

\paragraph{Insurance}
The Insurance dataset is another regression type dataset for predicting billed medical expenses \citep{lantz2019machine}. 
It is a relatively small dataset with 1338 observations. Besides the target variable, \textit{charges}, it contains 6 variables.

\begin{itemize}
    \item \textbf{Targets:} Charges
    \item \textbf{Features:}
    \begin{enumerate}
        \item \textit{Age}: The patient's age                            
        \item \textit{Sex}: Gender of the patient                        
        \item \textit{BMI}: The body-mass index of the patient      
        \item \textit{Children}: The number of children the patient has    
        \item \textit{Smoker}: Whether the patient is a smoker or not            
        \item \textit{Region}: The region where the patient lives, categorized into 4 sections, southeast, southwest, northwest and northeast
                  
    \end{enumerate}
\end{itemize}

\paragraph{Credit}
The credit card dataset \citep{dal2015adaptive} is a dataset used for fraud detection and hence highly unbalanced. It is comprised of 108754 observations and only 0.172\% of the observations are fraudulent. It is also used as a benchmark dataset by \citet{agarwal_neural_2021}.
The dataset is comprised of 30 numerical input variables in addition to the target variable which is a binary identifier whether there was financial fraud or not.
The meaning for most of the variables is not revealed due to confidentiality reasons \citep{dal2015adaptive}. All variables have already been transformed by principal component analysis. The few variables which are revealed are listed below.

\begin{itemize}
    \item \textbf{Targets:} Class $\rightarrow$ Fraud or not
    \item \textbf{Features:}
    \begin{enumerate}
        \item \textit{Time}: The seconds elapsed between each transaction and the first transaction in the dataset                           
        \item \textit{Amount}: The transaction amount                       
                  
    \end{enumerate}
\end{itemize}

\paragraph{FICO}
As a second dataset for a classification benchmark we use the FICO dataset \citep{FICO}. We use the \textit{Risk Performance} as the target variable and all other variables as features.
A detailed description of the features and their meaning is available at \citet{FICO}. The dataset is comprised of 10459 observations.

\paragraph{Preprocessing}
We implement the same preprocessing for all four dataset and closely follow \citet{gorishniy_revisiting_2021} in their preprocessing steps. Hence all numerical variables are scaled between -1 and 1, all categorical features are one-hot encoded and we implement quantile smoothing.
We use 5-fold cross-validation and report mean results as well as the standard deviations. For reproducability we use the sklearn \citep{pedregosa_scikit-learn_2011} Kfold function with a random state of 101 and shuffle equal to true for all datasets.

\subsection{Hyperparameter Tuning}
We implement light hyperparameter tuning over all models and closely follow \citet{agarwal_neural_2021} reported model structures.
Hence, we implement the Multilayer Perceptrons with 10 layers and 100 neurons each. Each layer uses the rectified linear unit (ReLU) activation functions and is followed by a 0.25 dropout layer. Similar to \citet{agarwal_neural_2021} we implement an early stopping with a patience of 100 epochs and find no overfitting for all models. Additionally, we reduce the learning rate by a factor of 0.995 once a plateau is reached and start with a learning rate of 1e-04 using the ADAM optimizer \citep{kingma2014adam}. All models are trained with an initial number of 1000 epochs, however, over all neural models, all datasets and all folds every model stopped training before reaching the maximum number of epochs. We iterate over different batch sizes and find that, similar to \citet{agarwal_neural_2021}, 1024 is the best batch size for the California Housing dataset and the FICO dataset. For the creditcard dataset we find a batchsize of 512 to perform best and for the Insurance dataset, which is the smallest of the four datasets, we find a smaller batchsize of 64 to perform best. We implement the Neural Additive Models \citep{agarwal_neural_2021} identically to the implementations described in \citet{agarwal_neural_2021}. Additionally, we experiment with the described ExU activation functions but find that simple ReLU activation functions outperform the ExU functions for 3 out of 4 datasets. For XGBoost we use the following hyperparameters: number of estimators=1000, maximum depth=25, eta=0.1, subsample=1.0, colsample by tree=0.95. Interestingly, XGBoost performs extremely well on 3 out of 4 datasets but is clearly outperformed by all neural approaches for the FICO dataset. For EBM we use the following hyperparameters: max leaves=10, learning rate=0.005, early stopping rounds=100. EBM perform very similar to XGBoost but outperform all models for the small Insurance dataset.

%\begin{figure*}
%         \centering
%         \includegraphics[width=\textwidth]{img/G%AM_Units.png}
%         \caption{The figure sketches how the %different types of approximations could %be implemented as neural networks.}
%         \label{fig:generalized_layers}
%\end{figure*}

%\begin{table}[H]
%\small
%\begin{tabular}{|l|l|}
%\hline
%\textbf{Feature Name} & \textbf{Description}                                         \\ \hline
%MedInc                & Median income of the area                                    \\ \hline
%HouseAge              & Median house age of the area                                 \\ \hline
%AveRooms              & Area-wise average number of rooms per household              \\ \hline
%AveBedrms             & Area-wise average number of bedrooms per household           \\ \hline
%Population            & Total number of people living in the area                    \\ \hline
%AveOccup              & Area-wise average number of people within a household \\ \hline
%Latitude              & Latitude of the center of the area                           \\ \hline
%Longitude             & Longitude of the center of the area                          \\ \hline
%\end{tabular}
%\end{table}

\end{document}